%% file: main_elsarticle-template-harv.tex
\documentclass[final,1p,authoryear]{elsarticle}

\usepackage{amsmath,amsfonts, amssymb}
\usepackage{algorithmic}
\usepackage{array}
\usepackage[utf8]{inputenc}
\usepackage[caption=false,font=normalsize,labelfont=sf,textfont=sf]{subfig}
\usepackage{textcomp}
\usepackage{stfloats}
\usepackage{url}
\usepackage{verbatim}
\usepackage{graphicx}
\usepackage[]{subfig}
\usepackage{balance}
\usepackage{stfloats}
\usepackage{hyperref}
\usepackage{lineno}
\usepackage{hyphenat}

\graphicspath{{images/}}
\renewcommand{\d}[1]{{\mbox{\boldmath$#1$}}}

\makeatletter
\def\ps@pprintTitle{%
  \let\@oddhead\@empty
  \let\@evenhead\@empty
  \let\@oddfoot\@empty
  \let\@evenfoot\@oddfoot
}
\makeatother

\begin{document}
\begin{frontmatter}



\title{Exploring Self-Attention for Crop-type Classification Explainability\tnoteref{t1}}

\tnotetext[t1]{This work is supported by the Munich Center for Machine Learning (MCML), by the German Federal Ministry of Education and Research (BMBF) in the framework of the international future AI lab "AI4EO -- Artificial Intelligence for Earth Observation: Reasoning, Uncertainties, Ethics and Beyond" (grant number: 01DD20001) and by German Federal Ministry for Economic Affairs and Climate Action in the framework of the "national center of excellence ML4Earth" (grant number: 50EE2201C). We thank Marc Rußwurm for his constructive feedback.}

\author[1,4]{Ivica Obadic\corref{corr_author}}
\ead{ivica.obadic@tum.de}

\author[2]{Ribana Roscher\fnref{fn_credit}}
\ead{ribana.roscher@uni-bonn.de}

\author[1,3]{Dario Augusto Borges Oliveira}
\ead{dario.oliveira@tum.de}

\author[1,4]{Xiao Xiang Zhu\corref{corr_author}}
\ead{xiaoxiang.zhu@tum.de}

\affiliation[1]{organization={Chair of Data Science in Earth Observation, Technical University of Munich},%Department and Organization
            addressline={Arcisstraße 21}, 
            city={Munich},
            postcode={80333}, 
            country={Germany}}

\affiliation[2]{organization={Forschungszentrum Jülich GmbH, Institute of Bio- and Geosciences, Plant Sciences},%Department and Organization
            addressline={Wilhelm-Johnen-Straße}, 
            city={Jülich},
            postcode={52428}, 
            country={Germany}}

\affiliation[3]{organization={School of Applied Mathematics, Getulio Vargas Foundation},%Department and Organization
            addressline={Praia de Botafogo, 190}, 
            city={Rio de Janeiro},
            postcode={22250-900}, 
            country={Brazil}}

\affiliation[4]{organization={Munich Center for Machine Learning (MCML)},%Department and Organization
            addressline={Arcisstraße 21}, 
            city={Munich},
            postcode={80333}, 
            country={Germany}}

\cortext[corr_author]{Corresponding authors.}
\fntext[fn_credit]{The major contribution of Ribana Roscher to this manuscript was during her time at the Chair of Data Science in Earth Observation, TUM and with IGG, Remote Sensing, University of Bonn.}

\input{abstract}

% \begin{highlights}
% \item An explainable machine learning framework is presented to improve the transparency of a state-of-the-art transformer encoder model for crop-type classification.
% \item Our framework relates the attention weights of a trained transformer encoder to domain knowledge about crop phenology to identify the critical dates and the phenological events for crop-type classification.
% \item The proposed sensitivity analysis of the attention weights reveals that the identified phenological events are relative to the set of crops considered during model training. 
% \end{highlights}
\end{frontmatter}

%\begin{linenumbers}
\vspace{1cm}
\input{introduction}
\input{related_work}

\input{methods}
\input{explainability_framework}
\input{experimental_setup}
\input{experiments}
\input{discussion}
\input{conclusion}




\bibliographystyle{elsarticle-harv}
\bibliography{references}

\input{appendix}

%\end{linenumbers}

\end{document}

%% file: abstract.tex
\begin{abstract}
% WHY is it relevant?
% Automated crop-type classification using Sentinel-2 satellite time series is essential to support agriculture monitoring. 
%Transformer encoder models became a promising approach for crop-type classification and their attention weights were explored to reveal the temporal importance assigned by these models. Yet, these interpretability insights come with questionable validity as it is uncertain how close the attention weights approximate the actual workings of the black-box models.
Transformer models have become a promising approach for crop-type classification. Although their attention weights can be used to understand the relevant time points for crop disambiguation, the validity of these insights depends on how closely the attention weights approximate the actual workings of these black-box models, which is not always clear.
% WHICH PROBLEM do we address?
%Using explainable machine learning to reveal the inner workings of these models is an important step towards improving stakeholders' trust and efficient agriculture monitoring. 
% HOW is our approach/analysis special, WHAT are we actually doing, and WHAT IS NEW
%In this paper, we unveil the explanatory power of the attention weights for crop-type classification by introducing a novel explainability framework that relates the attention weights to domain knowledge about crop phenology to identify the critical dates and the phenological events for crop disambiguation. Next, we use these insights to evaluate the quality of the attention weights explanations and further present a sensitivity analysis approach to understand better the attention capability for revealing crop-specific phenological events.
%In this paper, we introduce a novel explainability framework for crop-type classification that leverages the attention weights of transformer encoder models to reveal the temporal importance of different features for crop-type disambiguation.
In this paper, we introduce a novel explainability framework that systematically evaluates the explanatory power of the attention weights of a standard transformer encoder for crop-type classification.
Our framework first relates the attention weights to domain knowledge about crop phenology to interpret the salient dates and phenological events for the model predictions. Next, we evaluate whether these insights are critical for crop disambiguation and develop a sensitivity analysis approach to understand the capability of attention to reveal crop-specific phenological events.
%% EVALUATION, WHAT FOLLOWS, ETC.
%Our results show that attention patterns strongly relate to key dates, and consequently, to the critical phenological events for crop-type classification which might be relevant for improving stakeholder trust and optimizing agriculture monitoring processes. Additionally, the sensitivity analysis demonstrates the limitation of the attention weights to reveal the important events in the crop phenology as the unveiled phenological events depend on the other crops in the data considered during training.
Our results show that attention patterns strongly relate to key dates, which are often associated with critical phenological events for crop-type classification. %This finding is relevant for improving stakeholder trust and optimizing agriculture monitoring processes because it helps to understand how the model is making its predictions and to identify potential areas for improvement.
Further, the sensitivity analysis reveals the limited capability of the attention weights to characterize crop phenology as the identified phenological events depend on the other crops considered during training. This limitation highlights the relevance of future work towards the development of deep learning approaches capable of automatically learning the temporal vegetation dynamics for accurate crop disambiguation.%That is important to guide attention weights interpretability for crop-type classification, as it shows the attention weights are strongly anchored on data, and therefore prevent extending the interpretation to unseen crops in the dataset.
\end{abstract}

%% file: introduction.tex
\section{Introduction}
\noindent
\label{sec:introduction}
%% WHY
% \emph{WHY: First, answer the WHY question. Why is that relevant? Why should I be
% motivated to read the paper}
%\the\textwidth
Monitoring crop fields is a vital task for agriculture, and in the European Union (EU), it plays an essential role in the decision process for agricultural subsidization. 
%The authorities verify whether the crop type grown on an agricultural parcel matches the farmer's declaration and ensure  %In this context, the introduction of the Sentinel missions fostered the development of automated pipelines for efficient crop monitoring based on the freely available Sentinel satellite observations.
According to the audit report by the \cite{new-tech-in-agri-monitoring-4-2020}, several EU members already use machine learning algorithms trained on time series of Sentinel observations for crop-type classification and detection of various phenological events. 
%% WHICH PROBLEM
% \emph{WHICH PROBLEM: Second, explain WHICH problem you are solving/address to
% solve.}
In the last years, state-of-the-art approaches for crop-type classification were proposed using transformer encoder models \citep{RUWURM2020421,garnot2020satellite, garnot2022multi}. Notwithstanding the accurate crop maps produced by these models, their black-box nature disables a straightforward understanding of the model decision process. Yet, for improving trust and enabling broad adoption of such models for agricultural policy making, it is desirable to connect model decisions to common agricultural knowledge \citep{campos2020understanding}.

At the core of the transformer encoders is the self-attention mechanism which models the temporal dependencies in the data through the attention weights. They indicate the relevance of the sequence elements when creating functional, high-level feature representation \citep{vaswani2017attention}. Inspecting the attention weights became a popular interpretability approach used to understand the workings of the transformer models in natural language processing \citep{clark-etal-2019-bert}, image classification \citep{li2023does} or video action recognition \citep{meng2019interpretable}. Although the attention weights can be used to asses the temporal importance of the satellite observations \citep{RUWURM2020421, xu2021towards}, their potential is far from fully explored for uncovering other important aspects in crop monitoring, such as identification of phenological events. At the same time, the validity of the explanations based on attention weights is challenged in several studies which yield no clear consensus on whether the attention weights faithfully explain the model decisions \citep{bibal2022attention}. Therefore, with the growing number of black-box approaches for crop-type classification that rely on self-attention, it becomes essential to:
\begin{enumerate}
    \item explore the potential of attention weights to reveal the inner workings of these models, and
    \item unveil their explanatory power in the context of crop disambiguation. 
\end{enumerate}

In this paper, we tackle the above questions by applying a novel explainability framework that leverages the attention weights of a trained transformer encoder model to identify critical insights for agriculture monitoring. Next, we evaluate the relevance of the attention weights explanations for crop disambiguation and assess their potential for uncovering detailed events in crop phenology. In summary, our proposed methodology improves the transformer encoder model explainability for crop-type classification and uncovers its limitations with the following main contributions:
%% MAIN CONTRIBUTION & WHAT FOLLOWS FROM THAT
\begin{itemize}
  \item Identification of the relevant dates and phenological events for the transformer encoder model by relating the sparse attention patterns with domain knowledge about crop phenology.
  \item Quantitative evaluation of the attention weights explanations to assess whether the identified insights are crucial for crop disambiguation.
  \item Sensitivity analysis to better understand the capabilities of the attention weights for detecting important events in crop phenology.
\end{itemize}

Our findings show that the self-attention mechanism highlights the key dates for crop disambiguation, as it assigns high attention values to crops exhibiting distinct spectral reflectance features. Moreover, attention patterns provide insights into specific and relevant events in crop phenology, such as harvesting and growing, which play a critical role in effectively distinguishing between different crop types. However, it is important to note that our sensitivity analysis indicates that attention weights do not capture all significant events in crop phenology. The identified phenological events are conditioned on the presence of other crop types within the dataset, meaning that they are specifically relevant for disambiguating between different classes.

%% file: related_work.tex
\section{Related Work}
\noindent
\label{sec:related_work}
%The rising popularity of deep neural networks came together with a negative stamp of black\hyp{}box models, as they usually do not explicitly unveil the mechanisms to justify the predicted outcome. That led to the emergence of explainable machine learning which aims to improve trust in these models by revealing the learned patterns for their inference \citep{molnar2020interpretable}. A significant class of explainable machine learning methods consists of approaches based on sensitivity analysis \citep{ras2022explainable}. One of the benchmark methods in this class is Occlusion Sensitivity Maps \citep{zeiler2014visualizing}, which attributes high importance to image patches whose occlusion leads to a decrease in the model performance. This method is also popular in applications of explainable machine learning in environmental sciences. For example, \citep{kierdorf2020identifies} uses it to discover relevant pixels for whale identification and \citep{stomberg2022exploring} presents an approach based on the occlusion of latent space activations to explain wilderness characteristics in Sentinel-2 images.
%The workings of attention-based architectures such as transformers are commonly interpreted by using the attention weights \citep{meng2019interpretable, wang2016attention}.
% restructure it by saying what the approaches aim to explain and then higlight the difference with the proposed approach
The interpretability studies for crop\hyp{}type classification typically focus on understanding the importance of the temporal signal and identifying the relevant spectral bands. For example, \citep{vuolo2018much} shows that utilizing Sentinel-2 observations acquired within the growing season improves the accuracy of a random forest model. Further, \citep{campos2020understanding} propose a perturbation method that reveals high relevance of the Sentinel\hyp{}2 observations acquired in the summer months and the red and the near-infrared band for the predictions of a BiLSTM model. Similar findings are observed in \citep{xu2021towards} where feature importance analysis reveals that attention\hyp{}based deep learning approaches and a random forest model attribute high importance to the observation acquired in the summer months and to the shortwave infrared band. The growing number of crop-type classification models based on self-attention led to using the attention weights for understanding the temporal importance patterns assigned by these models \citep{RUWURM2020421,garnot2020satellite,garnot2020lightweight, xu2021towards}. These studies find that the attention weights are primarily concentrated on short and specific temporal intervals per crop type. Moreover, \citep{RUWURM2020421} discovers that the transformer encoder suppresses the cloudy observations and argues that this behavior derives from the sparse attention distribution that neglects the cloudy observations.  
%On the other hand, the detection of phenological events is usually tackled by time series analysis of vegetation indices \citep{new-tech-in-agri-monitoring-4-2020, hashemi2022assessing}.
At the same time, the usage of attention weights for model explanation is questioned in several natural language processing studies that investigate how close the attention weights approximate the inner model workings \citep{bibal2022attention}. For instance, \citep{jain-wallace-2019-attention} observes a weak correlation between the attention and gradient-based feature importance and the existence of adversarial attention distributions or \citep{meister2021sparse} shows that inducing sparsity in the attention distribution does not improve some of the interpretability benchmarks. Therefore, while the existing works demonstrate that attention-based explanations can offer valuable insights into the behavior of the transformer encoder model for crop-type classification, it is important to note that they do not directly elucidate the impact of attention weights on model predictions nor highlight the potential of these weights in identifying significant events in crop phenology. Addressing these crucial aspects is the primary objective of our explainability framework, which is comprehensively presented in this paper.
%\citep{serrano-smith-2019-attention} proposes alternative criteria that demonstrate that the prior work does not disprove the usage of attention as an explainability tool. 

%% file: methods.tex
\section{Methods}
\subsection{Self-Attention Mechanism}
\label{sec:self-attention-mechanism}
The self-attention mechanism \citep{vaswani2017attention} creates high-level feature representations for the sequence elements by modeling the temporal dependencies in the data. First, the input sequence $X \in R^{T \times d_{in}}$ is linearly projected into a query matrix $Q \in R^{T \times d_{k}} $, a key matrix $K \in R^{T \times d_{k}} $ and a value matrix $V \in R^{T \times d_{v}}$ with the following operations:
\begin{equation}
\label{eq:key_query_val_projections}
Q=X \theta_Q,\: K=X \theta_K \:, \: V=X \theta_V
\end{equation}
where $T$ is the length of the time series, $d_{in}$ is the embedding dimension of the sequence elements,  $d_{k}$ is the query and key embedding dimension, $d_{v}$ is the value embedding dimension and
$\theta_Q$, $\theta_K$, and $\theta_V$ are projection matrices optimized jointly with the usual model parameters during the training process.
%The $\theta$ matrices are optimized jointly with the usual model parameters during the training process. 
%This allows the model to automatically learn task-specific feature representations from the input data.
\\
Next, the "Scaled Dot\hyp{}Product Attention" combines the keys and the queries into attention weights with the following equation:
\begin{equation} 
\label{eq:attn_weights}
A = \operatorname{softmax}(\frac{QK^{T}}{\sqrt{d_k}})
\end{equation}
$A \in R^{T \times T}$ is a square matrix containing the attention weights which model the alignment between the queries and keys.
These weights are used as linear coefficients to relate the different positions in the sequence. Concretely, an entry $a_{ij}$ of the matrix $A$ models the influence of the $j$-th sequence element in creating the high-level feature representation $h_i$ for the $i$-th  sequence element with the following equation:
\begin{equation}
\label{eq:attn_embeddings}
    h_i = \sum_{j=1}^{T} a_{ij}\d{v_j}
\end{equation}
where $\d{v_j}$ is the $j$-th row in the value matrix $V$ representing the value embedding for the position $j$.

\citep{vaswani2017attention} also introduces the "Multi\hyp{}Head Attention" approach which on parallel projects non\hyp{}overlapping subspaces of the input data into queries, keys, and values and concatenates the output of the scaled dot-product attention on each projected version. Hence, the number of heads corresponds to the number of applied projections.

\subsection{Transformer Encoders}
\label{sec:transformer_encoder}
%General about transformers - why do they exist
Transformer encoders \citep{vaswani2017attention} are deep learning architectures composed of encoder layers that rely on the self-attention mechanism for modelling the temporal dependencies relevant for inference over sequential data. These layers further process the output of the self-attention mechanism through a residual block and a dense layer.
It is important to note that before the data is fed into the encoder layers, the transformer first adds positional encoding to the input in order to inform the model about the order of the sequence elements.

\subsection{Explanation Evaluation}
\label{sec:explanation_evaluation}
Evaluating how close an explanation method approximates the predictive behavior of a machine learning model is frequently tackled by measuring the change in the model predictions after the occlusion of the relevant features \citep{hedstrom2023quantus}. However, an occlusion during model evaluation can introduce out-of-distribution examples to the model and this raises the question of whether the decrease in the model performance can be attributed solely to the occluded features. To address this issue, \citep{hooker2019benchmark} proposes a framework to evaluate the explanation methods based on 1) the removal of the relevant features and 2) model retraining. First, a modified dataset is created where a fraction of the most important features indicated by the explanation method are removed. Next, the model is trained on the modified dataset and its accuracy is compared against the accuracy of the model trained on the entire dataset. Observing a drop in accuracy after the occlusion of a small number of supposedly relevant features indicates that the explanation method highlights the features relevant to the learning task. 

%% file: explainability_framework.tex
\section{Explainability Framework}
\noindent
\begin{figure}[t!]
\centering
\includegraphics[width=1.3\textwidth]
{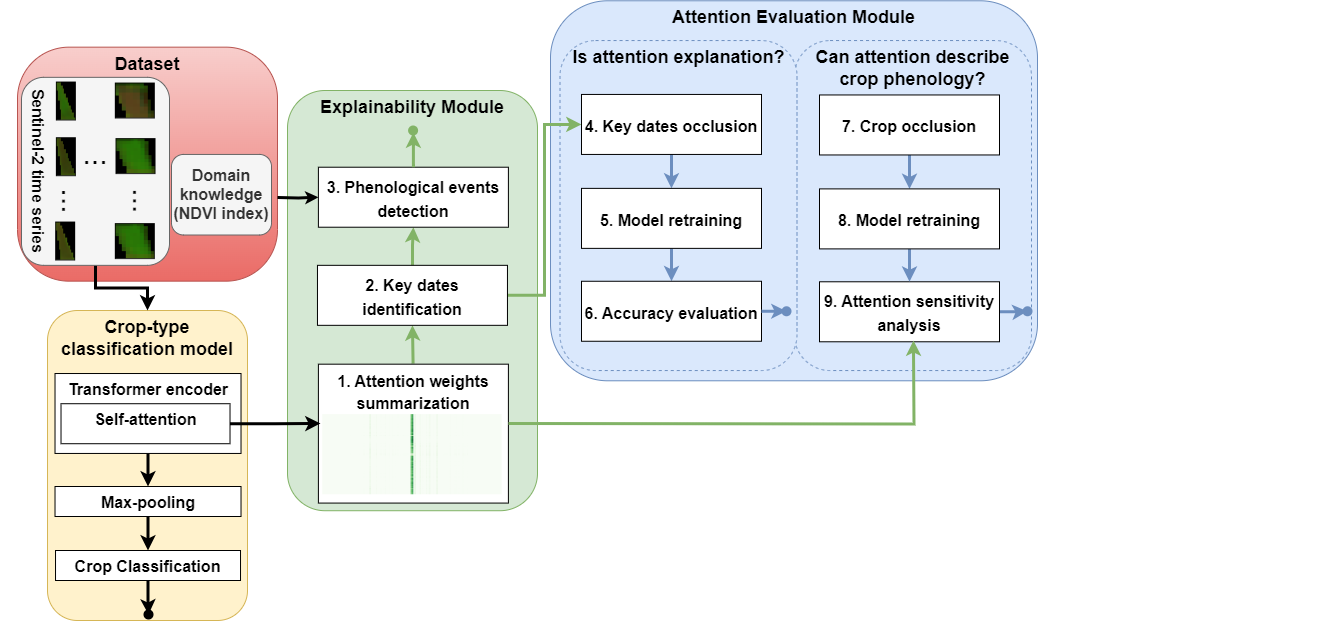}
\caption{Diagram of the explainability framework presented in this paper. The explainability module (depicted in green) processes the attention weights of a trained transformer encoder for crop-type classification (depicted in yellow) to identify the key dates and phenological events that support model predictions. Next, the attention evaluation module (depicted in blue) occludes the identified key dates to assess their importance for crop disambiguation and performs crop occlusion to test the attention capability to characterize crop phenology.}
\label{fig:explainability_framework_diagram}
\end{figure}
As discussed in Section \ref{sec:introduction}, although investigating the attention weights learned using self\hyp{}attention became a popular explainability approach, its potential is still largely unexplored for crop\hyp{}type classification. In this paper, we reveal the explanatory power of the attention weights for crop-type classification with the explainability framework visualized in Figure \ref{fig:explainability_framework_diagram}. It is based on the following two core modules: 1) the \textbf{explainability module} depicted in the green rectangle and the \textbf{attention evaluation module} depicted in the blue rectangle. First, the explainability module summarizes the attention weights of a trained transformer encoder model (depicted in the yellow rectangle) to identify the salient dates. Next, it relates the attention patterns on these dates with domain knowledge about the crop phenology extracted from the dataset (depicted in the red rectangle) to detect the phenological events for crop disambiguation. Finally, the attention evaluation module investigates whether the identified key dates are critical for the model decisions and unveils the capability of the attention weights to unambiguously reveal important events in the crop phenology.

\subsection{Crop-Type Classification Model}
\label{sec:crop_type_classification_model}
The goal of crop-type classification is to learn a mapping from time series of satellite images for an agricultural parcel to its crop type. For this purpose, we train the transformer encoder model proposed by \cite{RUWURM2020421}. This model first preprocesses the input by adding positional encoding to the sequence and linearly mapping the sequence embeddings to a higher dimension. Next, the sequence is processed through a series of transformer encoder layers described in Section \ref{sec:transformer_encoder}. The last encoder layer is followed by a max-pooling layer which extracts the highest feature at each time point and a fully connected classification layer that predicts the crop type.

\subsection{Explainability Module}
\label{sec:explainability_module}
The explainability module uses the attention weights to reveal the key dates for the predictions of the trained transformer encoder model. These findings are connected to domain knowledge about crop phenology to explain the phenological events relevant to crop-type classification. We propose the following steps to achieve this:
\subsubsection{Attention Weights Summarization}
As stated in Section \ref{sec:transformer_encoder}, the transformer encoder models the temporal dependencies in the data through the attention mechanism. 
%Therefore, in order to uncover the temporal importance assigned by the trained transfomer encoder model for crop-type classification, we analyze the resulting attention weights patterns.
A typical attention weights matrix for crop-type classification is illustrated at the bottom of the explainability module in Figure \ref{fig:explainability_framework_diagram}. It shows sparsely distributed attention weights, with the entire attention being assigned only to a few columns in the matrix. Based on the definition of the self-attention mechanism given in Section \ref{sec:self-attention-mechanism}, such pattern implies that the new embeddings consist only of the observations acquired on the dates that are highly attended. Therefore, we estimate the importance of date $d$ for classifying agricultural parcel $p$ as the average of the attention weights assigned to date $d$ with the following equation:
\begin{equation}
  \label{eq:obs_day_d_importance}
      \alpha_{d_p} = \frac{1}{T}\sum_{i=1}^{T}A_{{id}_{p}}
  \end{equation}
where $T$ is the number of temporal observations for parcel $p$ and $A_{{id}_p}$is the attention weight from date $d$ to date $i$ for the parcel $p$. 

Next, we reveal the importance that the model assigns to date $d$ for crop type $c$ with the following equation:
\begin{equation}
  \label{eq:obs_crop_type_importance}
       \alpha_{d_{c}} = \frac{1}{N_c}\sum_{p=1}^{N_c}  \alpha_{d_p}
\end{equation}
where $N_c$ is the number of agricultural parcels with the crop type $c$ in the test dataset.

\subsubsection{Key Dates Identification}
\label{sec:key_dates_learning}
We further propose to compute the overall date importance for the transformer encoder model with the following equation:
\begin{equation}
  \label{eq:obs_day_global_importance}
       \alpha_{d} = \frac{1}{N}\sum_{p=1}^{N}  \alpha_{d_p}
\end{equation}
where $N$ is the total number of field parcels in the test dataset. Consequently, as key dates, we consider the dates with the highest attention averages across the agricultural parcels in the test dataset.

\subsubsection{Phenological Events Detection}
\label{sec:phenological_events_discovery}
To uncover the phenological events to which the trained transformer encoder model points, we relate the attention temporal importance on the identified key dates to individual agricultural parcels and to domain knowledge about crop phenology expressed through the NDVI index. It is commonly used as a proxy for understanding crop phenology and is computed by relating the near-infrared (NIR) band and the visible red (RED) band with the following equation:
\begin{equation}
    \label{eq:ndvi_index}
    \text{NDVI} = \frac{\text{NIR} - \text{RED}}{\text{NIR} + \text{RED}}
\end{equation}
The NDVI index ranges from -1 to 1. Sparse vegetation is usually characterized by values below the threshold of 0.4. NDVI index higher than this threshold typically depicts moderate and dense vegetation \citep{pan2015mapping}.

\subsection{Attention Evaluation Module}
While our explainability module outputs a set of dates and phenological events based on the attention weights, the validity of using attention as a tool for model explanations discussed in Section \ref{sec:related_work} poses the question to which extent the model decisions can be attributed to these identified features. Further, it remains unclear whether the attention weights are capable to describe detailed events in the crop phenology. These two questions are answered with the approaches presented below. 
\subsubsection{Is Attention an Explanation?}
\label{sec:is_attention_explanation}
To empirically verify whether the attention weights explain the relevant features for crop disambiguation, we propose the following procedure inspired by the evaluation framework presented in Section \ref{sec:explanation_evaluation} that inspects drops in the model accuracy after the removal of observations acquired on certain dates from the dataset:
\begin{enumerate}
    \item The observation acquisition dates are ranked according to the importance for crop disambiguation computed with Eq. \ref{eq:obs_day_global_importance}.
    \item A new dataset is created in which observation acquired at the top-$t$ key dates are removed from the agricultural parcels, $t$ $\in$ $\{1, 3, 5, 10, 15,..., T\}$ (where $T$ is the total number of dates).
    \item The transformer encoder model is trained and evaluated multiple times with different initializations on the newly created datasets and its accuracy is compared with the model trained on the complete dataset.
\end{enumerate}
%Observing that the accuracy of the original model can be approximated only with observation acquired over a small subset of key dates, would imply that the attention weights correctly identify the relevant features for crop disambiguation used by the model.

\subsubsection{Can Attention Describe Crop Phenology?}
\label{sec:attn_sensitivity_analysis}
To understand whether the attention patterns are pointing to relevant phenological events regardless of the crops under consideration or if the transformer encoder model is actually considering only the phenological events relevant for crop disambiguation, we present an approach that reveals attention sensitivity to the presence of various crop type configurations in the dataset. 
%The former would imply that the attention weights can be used to describe the crop phenology while the latter would imply that they are optimized only to specific dates relevant for disambiguating the considered crops.
In the first step, we create new datasets in which one of the crops from the original dataset is removed. Next, we train a transformer encoder model from scratch on each new dataset and analyze the difference in the attention weights distribution. Concretely, we analyze the change in importance of date $d$ for classifying crop type $c$ after the occlusion of the crop-type $o$, $\delta_{d_{c,-o}}$, computed with the following equation:
%TODO: Should this equation be kept in the text?
\begin{equation}
    \label{eq:attn_temporal_importance_change}
    \delta_{d_{c,-o}} =  \frac{1}{N_c}\sum_{p=1}^{N_c}  (\alpha_{d_{p,-o}} - \alpha_{d_p})
\end{equation} 
where $N_c$ is the number of agricultural parcels with the crop type $c$ in the test dataset, $\alpha_{d_{p,-o}}$ is the importance of date $d$ for a parcel $p$ for the model trained on the dataset without crop-type $o$ and $\alpha_{d_p}$ is the importance of date $d$ for a parcel $p$ on the model trained with the complete set of crops. Both $\alpha_{d_{p,-o}}$ and $\alpha_{d_p}$ are computed with Eq. \ref{eq:obs_day_d_importance}.
\\
%The presented sensitivity analysis approach is designed with the following intuition: If attention patterns reveal the important phenological events per crop type, then the temporal importance assigned with the attention mechanism should not change depending on the set of classes under consideration.

%% file: experimental_setup.tex
\section{Experimental Setup}
\label{sec:experimental_setup}
\begin{figure}[h]
\centering
{\includegraphics[]{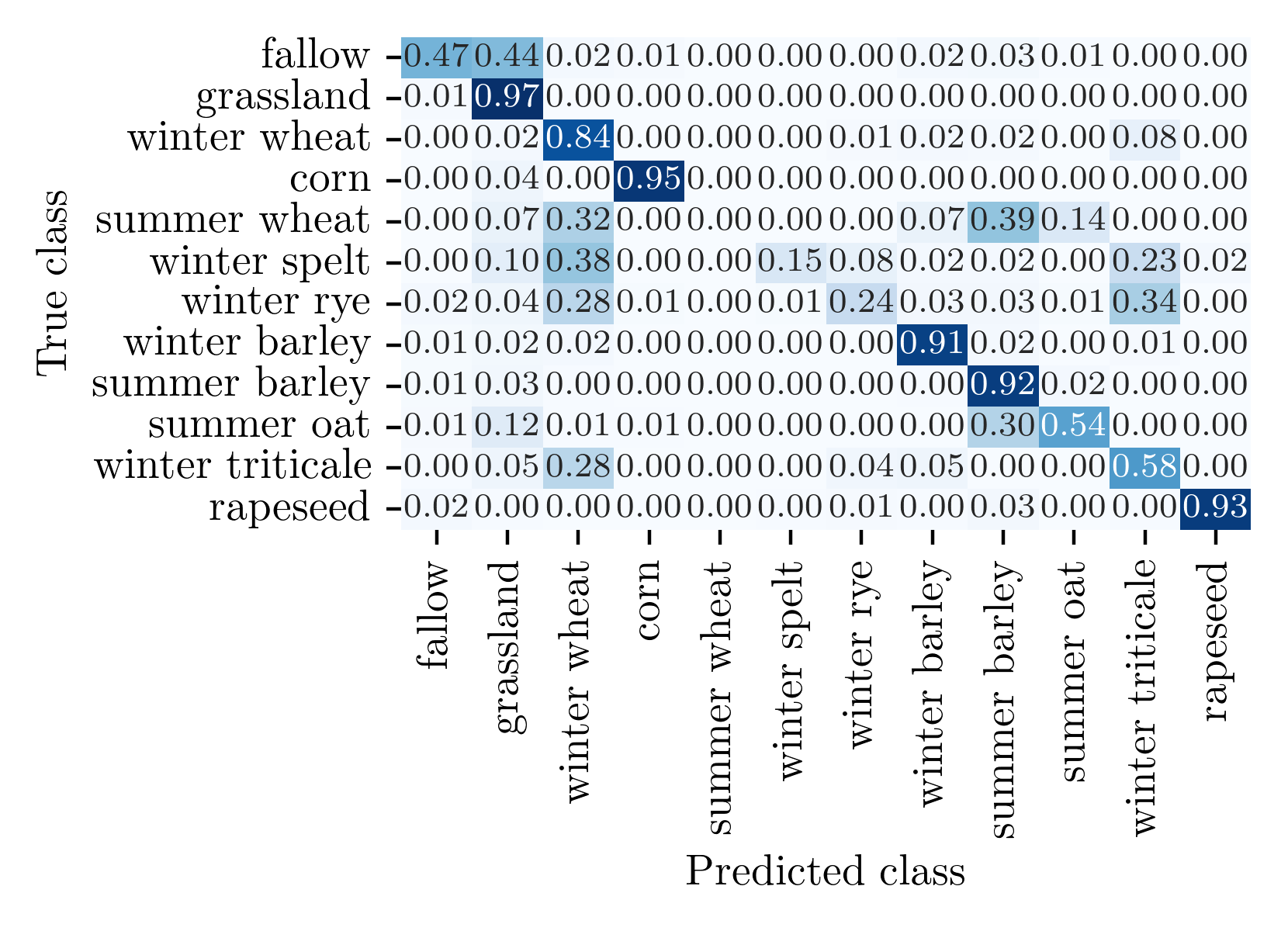}}
\caption{Confusion matrix of the transformer encoder model trained with the best hyperparameter combination. The model predicts with high accuracy the most frequent crops in the dataset as well as some of the less common crops like rapeseed.}
\label{fig:conf_matrix_best_model}
\end{figure}

\noindent
We performed our experiments on the BavarianCrops dataset \cite{RUWURM2020421} which contains agricultural parcels from Bavaria, Germany, represented with the pair $(x, y)$ where:
\begin{itemize}
    \item $x$ is a sequence of Sentinel-2 observations acquired over 2018 for the agricultural parcel. The image pixels within every observation were mean aggregated, thus, $x \in R^{T \times 13}$ where $T$ is the number of temporal observations. 
    \item $y$ is the crop grown on the agricultural parcel. We are using the dataset version consisting of 12 crops having imbalanced label distribution as the 5 most frequent crops cover 89\% of the agricultural parcels.
    %The label distribution is heavily imbalanced with the grassland crop type occurring on 59\% of agricultural parcels. It is followed by corn, winter wheat, winter barley and summer barley that appear on 13\%, 6\%, 6\% and 5\% of the agricultural parcels, respectively. 
\end{itemize}
As indicated in Section \ref{sec:crop_type_classification_model}, we train the transformer encoder model proposed by \cite{RUWURM2020421} and rely on the same dataset split and the training procedure, but we minimized the 
focal loss function \citep{lin2017focal} to tackle the class imbalance. Further, we perform different sequence preprocessing to consistently relate the attention weights to the temporal observations from the Sentinel-2 sequences. Concretely, we transform the input sequences to a fixed length of 144 (the maximum number of temporal observations per agricultural parcel) with the right padding and introduce a positional encoding based on the observation acquisition date to inform the model about the seasonality of the observations. The best model of the hyperparameter tuning described in Section \ref{sec:appendix_model_selection} is obtained with right padding, a positional encoding based on the observation acquisition date, 1 encoder layer, 1 attention head, and an embedding dimension of 128. It achieves a mean F1 score of 0.64 and average class accuracy of 0.62 which is similar to the results reported in \citep{RUWURM2020421}. The confusion matrix in figure \ref{fig:conf_matrix_best_model} shows that it correctly predicts the most frequent crops while low accuracy is observed for some of the less frequent crops. The explainability insights in the experiments were extracted based on this model that similar to the models trained for the sensitivity analysis, was initialized with a fixed seed of 0 to attribute the changes derived from class occlusion \footnote{Code: github.com/IvicaObadic/crop-type-classification-explainability}.

%% file: experiments.tex
\section{Attention Explainability Results}\label{sec:attention_explainability_results}
\subsection{Identifying Key Dates and Phenological Events for Crop Disambiguation}
\label{sec:exp_key_dates}
\begin{figure}[!h]
\centering
\includegraphics{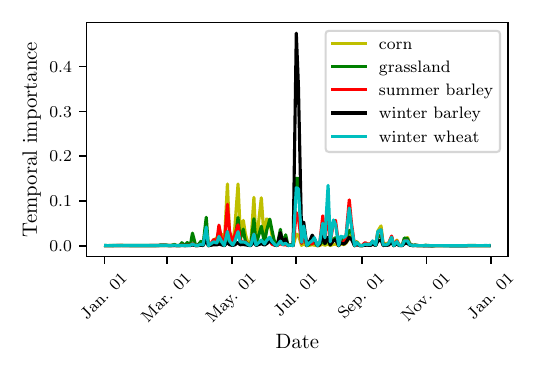}
\caption{Attention-based temporal importance computed with Eq. \ref{eq:obs_crop_type_importance} for the most frequent crops in the dataset. The model assigns high importance to the first days in July for predicting winter barley, to several dates in the Spring for predicting corn and to multiple dates throughout spring and summer for predicting grassland.}
\label{fig:attn_weights_over_time}
\end{figure}

As stated in Section \ref{sec:explainability_module}, we uncover the temporal importance assigned by the attention weights based on the assumption for sparse attention distribution. Figure \ref{fig:attn_weights_over_time} shows that the different crops present distinct attention patterns with high importance being assigned to observations acquired at a few specific dates per crop. Namely, the winter barley presents high attention only for the observations acquired at the beginning of July, the corn agricultural parcels are mostly attended at the observations acquired in late spring, and the attention for grassland spreads throughout the spring and summer months. Finally, the attention to observations acquired outside the spring and summer months in 2018 is typically sparse.

%\subsection{Connecting Attention Weights to Crop Phenology on the Key Dates for Crop Disambiguation}

\begin{figure*}[t!]
    \centering
    \subfloat[Winter barley parcel]{
        \includegraphics[width=0.5\textwidth]{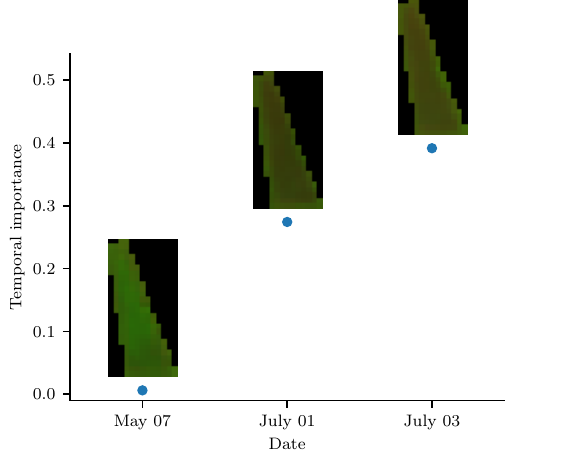}
        \label{fig:attn_over_time_winter_barley_parcel}}
    \subfloat[Corn parcel]{
        \includegraphics[width=0.5\textwidth]{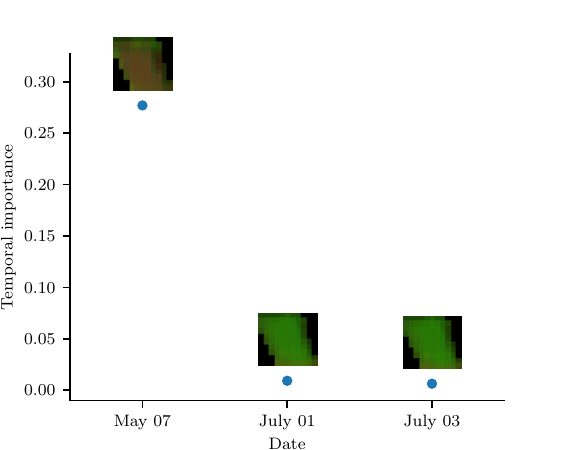}
        \label{fig:attn_over_time_corn_parcel}}
     
    \caption{Sentinel-2 observations and their temporal importance assigned by the model during the top-3 key dates for two example agricultural parcels. The temporal attention importance is calculated with Eq. \ref{eq:obs_day_d_importance} and is indicated by a dot for each date. The parcels are visualized with the combination of short-wave infrared (B11), Near Infrared (B8), and the blue band (B2) which highlights healthy vegetation in green and the sparsely vegetated areas in brown. The high importance of the observations acquired in July indicates that the harvesting event is crucial for winter barley prediction while the high importance of the observation acquired in May on the right plot highlights the importance of the growing event for corn prediction. The Appendix shows further examples of highly attended agricultural parcels in Figure \ref{fig:highest_attended_parcels_top_attn_dates}.}
    \label{fig:attn_over_time_parcels_visualization}
\end{figure*}

\begin{figure}[t]
     %NDVI vs attention 
     \subfloat[May 07]{
        \centering
        \includegraphics{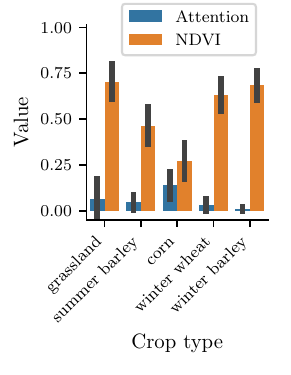}
        \label{fig:ndvi_vs_attention_05_07}}
    \subfloat[July 01]{
        \centering
        \includegraphics{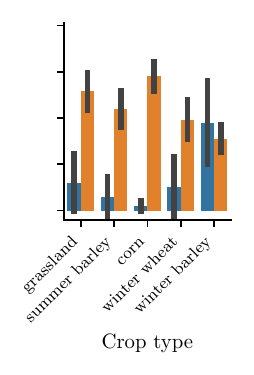}
        \label{fig:ndvi_vs_attention_07_01}}
    % \hfill
    \subfloat[July 03]{
        \centering
        \includegraphics{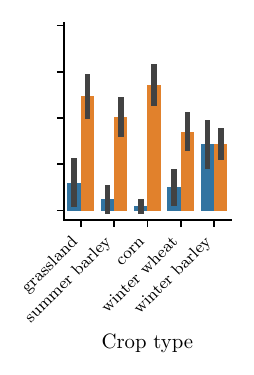}
        \label{fig:ndvi_vs_attention_07_03}}
    % \hfill

    \caption{The relation between the attention\hyp{}based temporal importance (Eq. \ref{eq:obs_day_global_importance}) and the average NDVI index per crop type on the top-3 key dates. The error bars show the standard deviation of the attention\hyp{}based temporal importance and NDVI index on each date. The attention mechanism assigns high importance to the crops displaying unique phenology on each key date.}
    \label{fig:attn_vs_NDVI_on_key_dates}
\end{figure}
%The above results show that the attention weights focus on the key dates in the time series in which the crops can be disambiguated based on the Sentinel-2 reflectances.
The above results show that the attention weights focus on a narrow set of dates per crop. In Section \ref{sec:phenological_events_discovery} we propose to detect the critical phenological events on these dates by relating the attention weights to individual agricultural parcels and to the NDVI index.
In Figure  \ref{fig:attn_over_time_parcels_visualization}, we show examples from the evolvement of winter barley and corn agricultural parcels and the corresponding attention-based temporal importance on the top-3 key dates. The winter barley example (Figure \ref{fig:attn_over_time_winter_barley_parcel}) presents green observation on May 07, indicating healthy vegetation, which has importance close to zero. On the other hand, the model assigns high importance to observations acquired at the beginning of July consisting mostly of brown pixels. Hence, the evolution of this parcel from healthy vegetation in spring to sparse vegetation in July points to the relevance of the harvesting event for predicting this agricultural parcel as winter barley.
The other example for the corn agricultural parcel (Figure \ref{fig:attn_over_time_corn_parcel}) illustrates that the model assigns high attention values to the observation acquired on May 07 which also mostly consists of brown pixels while the observations acquired in July displaying healthy vegetation are not attended. This pattern suggests that the features shown around the growing time influence the corn prediction.

To verify that the attention weights capture the above phenological events globally,  in Figure \ref{fig:attn_vs_NDVI_on_key_dates} we visualize the relationship between the average NDVI index and the temporal importance per crop type.
The association for May 07 (Figure \ref{fig:ndvi_vs_attention_05_07}) shows that attention highlights corn which is the crop having the most distinct phenology whose average NDVI index indicates sparse vegetation. Similarly, figures \ref{fig:ndvi_vs_attention_07_01} and \ref{fig:ndvi_vs_attention_07_03} show that for July 01 and July 03, the attention is focused on winter barley that again has a distinct NDVI index from the other crops pointing to sparse vegetation. 
In summary, from these results, we can conclude that the attention weights can be used to identify unique phenological events such as growing or harvesting that discriminate the crops on the dates with high importance for the model. 

% From the findings in this section we can be conclude that the attention weights guide the model to discriminate the crops not based on their spectral reflectance properties during their healthy vegetation, but rather by the difference in the stages of the crop phenology at specific time points.

\subsection{Does Attention Identify Relevant Features?}

\begin{figure}[h!]
\centering
\includegraphics{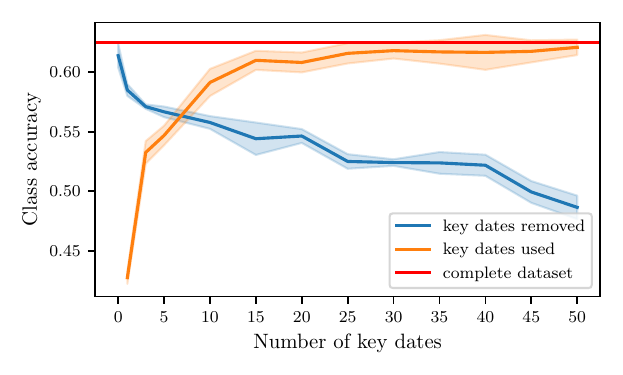}
\caption{Quantitative evaluation of attention weights explanations. This plot shows the average class accuracy and the standard deviation of multiple transformer encoder model training runs after the removal of the key dates (blue curve) and only by using the key dates (orange curve). The red curve illustrates the accuracy of the reference model trained on the entire dataset. The relevance of the first few dates identified with attention is demonstrated by a) a sharp decrease in the model accuracy after their removal (blue curve) and b) the sharp increase in model accuracy when these dates are incrementally added to the empty dataset (orange curve). The high percentage of the recovered accuracy after the removal of a large number of dates also shows that the dates not ranked among the first few most relevant also carry valuable information for crop disambiguation.
%These results indicate that the approach proposed in Section \ref{sec:key_dates_learning} can be used to identify the key dates in the year for accurate classification.
}
\label{fig:accuracy_top_N_dates}
\end{figure}
To assess whether the above-identified key dates and phenological events are critical for crop disambiguation, we evaluate the quality of the attention weights explanations with the procedure described in Section \ref{sec:is_attention_explanation} that is based on the removal of the dates with the highest attention averages and model retraining. The results of this ablation study are shown in Figure \ref{fig:accuracy_top_N_dates}. The blue curve shows that a sharp decrease in the model performance occurs with the incremental removal of the first 5 dates as the class accuracy drops by 5\% compared to the reference model trained on the entire dataset. The removal of the subsequent dates results in a steadier decrease in accuracy and the next larger gap by 2\% is observed after the removal of the 25th date. Yet, the accuracy of this model still amounts to 85\% of the accuracy of the reference model and subsequent removal of dates still results in models that approximate a large percentage of the accuracy of the reference model. To better understand these results, on the orange curve, we show the accuracy of the transformer encoder model trained on datasets consisting only of the dates with the highest attention averages. This curve is characterized by the sharpest increase in accuracy when only the first 3 dates are used for model training followed by a steadier increase until the first 15 dates are included which is sufficient for approximating the accuracy of the reference model. In summary, from this ablation study, we conclude that sparse attention weights identify a narrow set of dates in which the crops display unique phenological events critical for disambiguation. Nevertheless, the high percentage of the recovered accuracy of the reference model after the removal of the 20 most relevant dates also indicates the existence of multiple sparse sets of critical dates for crop disambiguation.

\subsection{Can Attention Describe Detailed Crop Phenology?}
\label{sec:exp_attn_sensitivity_analysis}
\begin{figure*}[t]
    \subfloat[Grassland occlusion]{
        \centering
        \includegraphics[]{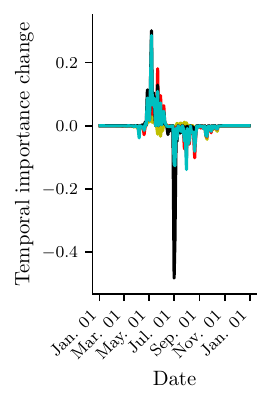}
        \label{fig:grassland_occlusion_attn_weights_change}}
    \subfloat[Winter barley occlusion]{
        \centering
        \includegraphics[]{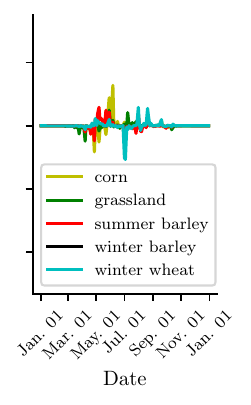}
        \label{fig:winter_barley_occlusion_attn_weights_change}}
    \subfloat[Corn occlusion]{
        \centering
        \includegraphics[]{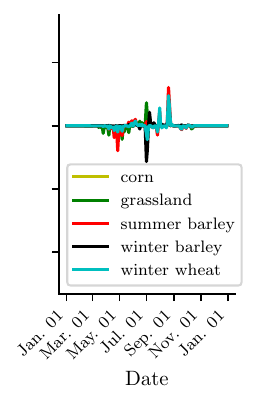}
        \label{fig:corn_occlusion_attn_weights_change}}
     
    \caption{Change in attention temporal patterns computed with Eq. \ref{eq:attn_temporal_importance_change} after the occlusion of grassland (Figure \ref{fig:grassland_occlusion_attn_weights_change}), winter barley (Figure \ref{fig:winter_barley_occlusion_attn_weights_change}) and corn (Figure \ref{fig:corn_occlusion_attn_weights_change}). 
    The temporal importance of a crop changes depending on the set of crops considered during model training. 
    %which implies that the phenological events discovered in the explainability module are conditioned on the presence of the other crops in the dataset.
    }
    \label{fig:attn_occlusion_change}
\end{figure*}

Using our findings about the identified critical set of dates and phenological events for crop disambiguation, we further investigate whether the attention weights are capable of uncovering these events regardless of the other crops present in the dataset. 
For this purpose, we perform the attention sensitivity analysis approach presented in Section \ref{sec:attn_sensitivity_analysis}. The results are presented in figure \ref{fig:attn_occlusion_change} which visualizes the change in the temporal importance for the models trained after removing grassland, winter barley and corn crops. Figure \ref{fig:grassland_occlusion_attn_weights_change} shows that occluding grassland drastically changes the resulting attention weights for the other crops. For example, the temporal importance for winter barley is shifted from the observations acquired in July (shown in Figure \ref{fig:attn_weights_over_time}) to the observations acquired in late April and May. This implies that the features shown around the harvesting time are not anymore relevant for winter barley discrimination when grassland is removed from the dataset.
%A similar pattern is observed for the winter wheat crop whose attention importance is also assigned to observations acquired in the spring months.
Although less symbolic when compared to the grassland occlusion, the attention weights also change after the occlusion of winter barley (Figure \ref{fig:winter_barley_occlusion_attn_weights_change}) and corn (Figure \ref{fig:corn_occlusion_attn_weights_change}). Consequently, our analysis shows that the attention weights do not independently model the behavior of the individual classes or have a general link to every relevant phenological event. Conversely, the interpretation of the attention weights is conditional to the classes in the dataset and focuses on events that help to disambiguate them.

%% file: discussion.tex
\section{Discussion}
The ability of attention weights to faithfully explain the workings of deep learning models is a subject of ongoing study and debate \citep{bibal2022attention}. The current claims are not settled, as the results depend on the employed model architecture, the learning task, and the evaluation criteria. With the increasing number of works that use self-attention to extract insights about model workings for crop-type classification, we propose to systematically investigate the explainability capabilities of attention weights in the context of agriculture monitoring to shed light on this debate.

%The power of the attention weights to faithfully explain the workings of the deep learning models is a subject of an ongoing study and debate \citep{bibal2022attention}. The current claims are contradicting as the results depend on the employed model architecture, the learning task, and the evaluation criteria. With the increasing number of works that rely on self-attention to extract insights about model workings for crop-type classification, it also becomes necessary to systematically investigate the explainability capabilities of attention weights in the context of agriculture monitoring. 
For this purpose, we implemented the evaluation procedure based on the remove and retrain framework described in Section \ref{sec:explanation_evaluation}. 
%Our results show that the largest decrease in model accuracy occurs after the removal of the top-5 most important dates identified with attention. The removal of the subsequent dates results in a smaller and steadier decrease in the model accuracy. 
As discussed by \citep{hooker2019benchmark}, observing a drop in accuracy with this procedure signals that identified features were informative to the model while a constant accuracy after removal and retraining might point to redundant input features. Consequently, the evaluation results presented in Figure \ref{fig:accuracy_top_N_dates} demonstrate that attention weights identified one sparse set of dates crucial for crop disambiguation. This finding aligns with the claims that the attention weights provide \textbf{an explanation} rather than \textbf{the explanation} to the model workings \citep{wiegreffe2019attention}, as the high percentage of retained accuracy of the reference model after the dates removal points to the existence of multiple sets of critical dates. Moreover, our results show that the transformer encoder model is capable of learning multiple plausible attention weights distributions that lead to similar model predictions. 

Although sparse attention weights can be empirically seen as an explanation for the model's decisions, they do not provide understandable agronomical insights on their own. Therefore, we follow the guidelines provided by \citet{roscher2020explainable} to extract scientific insights relevant to agriculture monitoring by introducing domain knowledge about crop phenology. This information has already been used to improve crop-type classification results in \citet{belgiu2021phenology}, and we use it to explain crop disambiguation patterns in terms of phenological events such as harvesting and growing. These insights can be relevant for cost reduction and schedule planning when acquiring very-high-resolution aerial images for monitoring small agricultural parcels \citep{new-tech-in-agri-monitoring-4-2020}. For example, we show that acquiring observations on a few dates throughout the year in which the crops are in different stages of their phenology is sufficient for accurate disambiguation. Additionally, the sensitivity analysis results in Section \ref{sec:exp_attn_sensitivity_analysis} show that identified phenological events are conditioned on the set of classes used to train the model. This implies that while the multi-temporal Sentinel-2 observations offer detailed insights into the crop phenology, the standard transformer encoder model does not utilize this information to disambiguate crops based on their specific vegetation dynamics throughout the vegetation season.

%% file: conclusion.tex
\section{Conclusion}
\noindent
In this paper, we presented an explainability framework that inspects the attention weights of a trained transformer encoder model to extract insights from the model workings for crop-type classification. We further present a quantitative evaluation procedure to systematically assess the validity of the attention weights as an explanation for the model's predictions in this context.

Our framework demonstrates that attention weights reveal that a much smaller set of key dates throughout the year is sufficient for accurate crop-type classification. By linking the attention patterns with agricultural domain knowledge, we show that the attention assigns high importance to distinct phenological events that discriminate between the classes at the key dates, such as the harvesting event for winter barley and the growing event for corn.

Our sensitivity analysis approach showed that the transformer encoder model does not comprehensively characterize crop-specific phenological cycles. Instead, it only highlights important events that are relevant for disambiguating the different crop classes considered during model training. In future work, we plan to address this limitation by designing a multi-layer transformer encoder architecture that introduces local window attention layers. The local attention is capable of identifying phenological events that span over consecutive time steps, allowing the model to capture long-term information about crop phenology and classify crops based on their specific temporal evolution.

%% file: appendix.tex
\appendix
 
\section{Model Selection and Hyperparameter Tuning}
\label{sec:appendix_model_selection}
\begin{figure}[h]
\centering
{\includegraphics{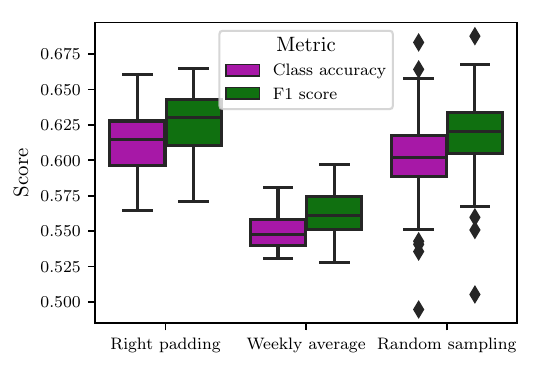}}
\caption{Hyperparameter tuning results per sequence aggregation approach. The right padding and random sampling outperform the weekly average aggregation approach.}
\label{fig:classification_results}
\end{figure}

We perform simple hyperparameter tuning for the transformer encoder parameters to select the model for the explainability analysis and compare the sequence preprocessing introduced in Section \ref{sec:experimental_setup} based on right padding with the random sampling approach and fixed positional encoding used in \citep{RUWURM2020421}. 
%where the input sequences are reduced to a standard length by randomly sampling 70 observations for every agricultural parcel and a fixed positional encoding is added based on the order in the sampled sequence.
We additionally evaluate a preprocessing that averages the observations for an agricultural parcel acquired within a calendar week. 
%We used the dataset split into training, validation and test set provided in \citep{RUWURM2020421}.
The results visualized in Figure \ref{fig:classification_results} illustrate that the right padding and random sampling approaches produce similar results with the right padding having higher classification scores on average. Both of these approaches significantly outperform the weekly average aggregation. Further, Table \ref{tab:right_padding_best_models} shows the top-3 hyperparameter configurations for the right padding sequence aggregation.
These results illustrate that the different hyperparameter configurations result in similar model performance.
%These results show that introducing sequence aggregation that allows us to consistently relate the attention weights to the temporal observations leads to a slight improvement in the classification results compared with the random sampling approach.

\begin{table}[!h]
\centering
\caption{Hyperparameters yielding best results for the right padding sequence aggregation approach.}
\label{tab:right_padding_best_models}
\begin{tabular}{|l|c|c|c|c|c|}
\hline
\textbf{Pos. enc}               & \textbf{Layers} & \textbf{Heads} & \textbf{Emb. dim} & \textbf{F1 score} \\ \hline
obs. acq.                       & 1                    & 1                   & 128               & 0.67                       \\ \hline
\multicolumn{1}{|c|}{seq. ord.} & 1                    & 1                   & 128               & 0.67                        \\ \hline
obs. acq.                       & 3                    & 1                   & 128               & 0.67                      \\ \hline
% obs. acq.                       & 2                    & 1                   & 128               & 0.66                         \\ \hline
% \multicolumn{1}{|c|}{obs. acq.} & 2                    & 2                   & 128               & 0.66                       \\ \hline
\end{tabular}
\end{table}

\section{Crop-Type Classification Results}
The classification results of the different models used in the attention explainability analysis are provided in Table \ref{tab:models_classification_results}. As indicated in Section \ref{sec:experimental_setup}, these models were trained with the best configuration found from the hyperparameter tuning and by setting a fixed seed of 0. The model trained on the entire dataset produces an accuracy of 0.62 and F1 score of 0.64, as shown in the first row of Table \ref{tab:models_classification_results}. %Using only the first key date is sufficient for reaching 71\% of the accuracy of the model trained on the entire dataset. Adding the following two key dates significantly improves the model accuracy to 86\% of the accuracy of the entire dataset.
The next row shows that in comparison with the overall accuracy of the model trained on the entire dataset, the occlusion of the grassland parcels leads to better overall discrimination for the remaining crops. On the other hand, the occlusion of the corn and winter barley leads to a decrease in the overall accuracy of the remaining crops. 
%In order to understand the influence of the crop occlusion on the accuracy of the remaining crops, in Figure \ref{fig:acc_change_after_occlusion} we visualize the change in the classification accuracy of the individual crops for the models trained on the datasets resulting after crop occlusion. Concretely, Figure \ref{fig:acc_change_grassland_occlusion} shows that occluding grassland leads to a drastic increase in the accuracy of some of the less frequent crops such as fallow (35\%) or summer oat (28\%). At the same time, it also leads to a 16\% decrease in accuracy for summer barley which is one of the most frequent crops in the data. Hence, these results point out that the large magnitude of change in the attention weights patterns after the grassland occlusion (shown in Figure \ref{fig:grassland_occlusion_attn_weights_change}) directly impacts the accuracy for some of the remaining crops. Furthermore, Figure \ref{fig:acc_change_winter barley_occlusion} shows that occluding the winter barley most significantly impacts the classification results of the other winter crops whereas the corn occlusion has little impact in the classification results for the other crops (figure \ref{fig:acc_change_corn_occlusion}).

\begin{table}[!h]
\centering
\caption{Results for the different models used in the proposed explainability framework.}
\label{tab:models_classification_results}
    \begin{tabular}{|c|c|c|}
        \hline
        \textbf{Dataset}        & \textbf{F1 score} & \textbf{Class accuracy} \\ \hline
        Entire Dataset          & 0.64              & 0.62                    \\ \hline
        %Top-1 key date          & 0.43              & 0.44                    \\ \hline
        %Top-3 key dates         & 0.54              & 0.55                   \\ \hline
        % Top-5 key dates         & 0.56              & 0.57                    \\ \hline
        Grassland occlusion     & 0.67              & 0.66                    \\ \hline
        Winter Barley occlusion & 0.59              & 0.58                    \\ \hline
        Corn occlusion          & 0.62              & 0.60                    \\ \hline
    \end{tabular}
\end{table}

\section{Agricultural Domain Knowledge}
\begin{figure}[h!]
\centering
{\includegraphics{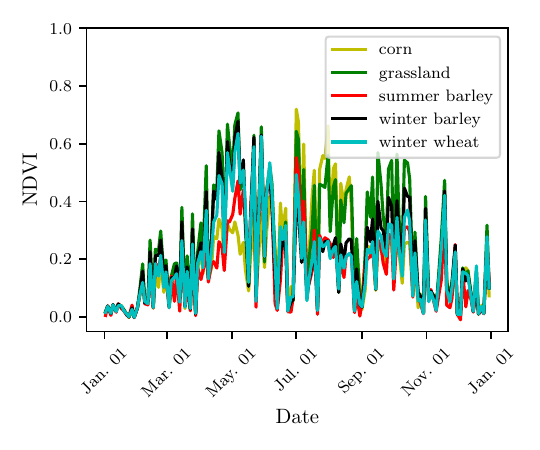}}
\caption{
Average NDVI index over time for the 5 most frequent crops in the dataset.}
\label{fig:ndvi_over_time}
\end{figure}

The NDVI temporal dynamics visualized in Figure \ref{fig:ndvi_over_time} shows that the winter crops are characterized by an earlier peak in dense vegetation in Spring and by sparser vegetation at the beginning of July, which indicates the harvesting event for these crops. This is in contrast to corn, whose NDVI index points to growth in May and harvesting in September. Summer barley has similar phenology as corn, with the distinction that it grows earlier and is harvested earlier. On the other hand, grassland phenology is characterized by dense vegetation throughout the entire year.

\begin{figure}[ht!]
\centering
    \subfloat[July 01]{
        \includegraphics[width=0.33\textwidth]{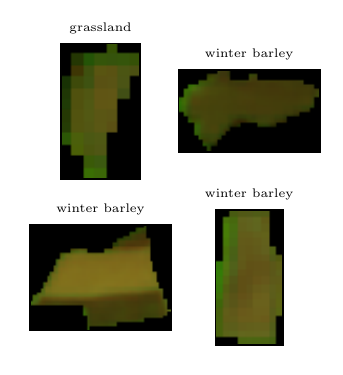}
        \label{fig:field_parcels_07_01}}
     \subfloat[July 03]{
        \centering
        \includegraphics[width=0.33\textwidth]{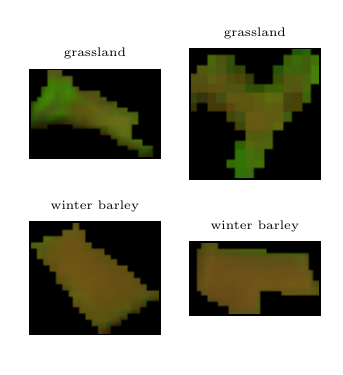}
        \label{fig:field_parcels_07_03}}
    \subfloat[May 07]{
        \centering
        \includegraphics[width=0.33\textwidth]{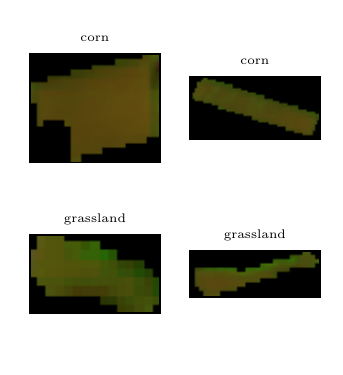}
        \label{fig:field_parcels_05_07}}
    \caption{The 4 highest attended parcels and their respective crop type on the top-3 key dates. %As in Figure \ref{fig:attn_over_time_parcels_visualization}, the parcels are visualized with the combination of short-wave infrared (B11), Near Infrared (B8) and the blue band (B2).% 
    The highest attended parcels mostly consist of brown pixels which indicate the relevance of the harvesting event on July 01 and July 03 and the relevance of the growing event on May 07 for crop disambiguation.}
    \label{fig:highest_attended_parcels_top_attn_dates}
 \end{figure}

Figure \ref{fig:highest_attended_parcels_top_attn_dates} shows the four parcels with highest attention according to Eq. \ref{eq:obs_day_d_importance} on the top-3 key dates for crop-type classifications. These parcels also consist mostly of brown pixels, confirming that the harvesting features displayed on July 01 and July 03 and the growing feature displayed on May 07 are critical for crop disambiguation.

